\documentclass[preprint,12pt]{elsarticle}

 \usepackage{epsfig}
\usepackage{subfigure}
\usepackage{amssymb,amsmath}
\def\argmax{\mathop{\rm arg\ max}\limits}

\newtheorem{theorem}{Theorem}[subsection]

\newtheorem{definition}[theorem]{Definition}
\newtheorem{example}[theorem]{Example}
\newtheorem{remark}[theorem]{Remark}

\begin{document}

\begin{frontmatter}



\title{Automated Query Learning with Wikipedia and Genetic Programming}

%%\title{Combining Wikipedia-semantics and Genetic Programming in Automatic Query Learning}


\author{Pekka Malo\corref{cor1}}
\cortext[cor1]{Corresponding author: Pekka Malo}
\ead{pekka.malo@aalto.fi}
\author{Pyry Siitari}
\ead{pyry.siitari@aalto.fi}
\author{Ankur Sinha}
\ead{ankur.sinha@aalto.fi}
\address{Aalto University School of Economics \\
 P.O. Box 21210, FI-00076 AALTO, FINLAND}

\begin{abstract}
Most of the existing information retrieval systems are based on bag of words model and are not equipped with common world knowledge. Work has been done towards improving the efficiency of such systems by using intelligent algorithms to generate search queries, however, not much research has been done in the direction of incorporating human-and-society level knowledge in the queries. This paper is one of the first attempts where such  information is incorporated into the search queries using Wikipedia semantics. The paper presents an essential shift from conventional token based queries to concept based queries, leading to an enhanced efficiency of information retrieval systems. To efficiently handle the automated query learning problem, we propose Wikipedia-based Evolutionary Semantics (Wiki-ES) framework where concept based queries are learnt using a co-evolving evolutionary procedure. Learning concept based queries using an intelligent evolutionary procedure yields significant improvement in performance which is shown through an extensive study using Reuters newswire documents. Comparison of the proposed framework is performed with other information retrieval systems. Concept based approach has also been implemented on other information retrieval systems to justify the effectiveness of a transition from token based queries to concept based queries.
\end{abstract}


\end{frontmatter}

% \linenumbers

\section{Introduction}\label{sec:introduction}

A central challenge in building expert systems for information retrieval (IR) is to provide them with common world knowledge. As succinctly put by Hendler and Feigenbaum~\cite{hendler01}, in order to build any system with ``significant levels of computational intelligence, we need significant bodies of knowledge in knowledge bases''. That is, if a system is expected to understand the general semantics in text, closer to the way human brains do, then it should have access to the extensive background knowledge that people use while interpreting concepts (units of knowledge) and their dependencies. Of course, statistical methods and natural language processing can be used to extract semantics from text or data, but the ability of text collections to convey human and society-level semantics is quite limited~\cite{zhuge10}. Currently, there is an ongoing quest to find new ways of integrating semantic knowledge into document modelling without time-consuming engineering. One of the emerging trends is to use socially developed resources of semantic information. 

In this paper, we consider the use of Wikipedia as a source of common world knowledge for an automated query learning system. The purpose is to assist users to express their information needs as queries which are written in terms of Wikipedia's concepts instead of word tokens. The proposed system extends the Inductive Query By Example (IQBE) paradigm of Smith and Smith~\cite{smith97} and Chen et al.~\cite{chen98} by incorporating human-level semantics using Wikipedia. The underlying principle of IQBE is quite simple: assume that a user provides a small collection of relevant (and irrelevant) example documents, the task is to learn a query based on those documents. The learnt query is then used to filter relevant documents from a newstream or document database according to the topic definition implied by the sample collection. The approach proposed in this paper uses concept-relatedness information contained in Wikipedia's link-structure to learn semantic queries using an intelligent co-evolutionary procedure. This transition from an ordinary boolean query~\cite{rijsbergen79} to a semantified query is necessary for integrating human and society-level semantic information into the information retrieval (IR) system. The integration of concept-based knowledge into the IR system enables it to detect the relevance of a document based on concepts and not just words. It also allows the system to identify those documents as relevant which contain concepts closely related to the query concepts. The paper contributes towards construction of an IR framework where Wikipedia-concept based queries are learnt using a co-evolving genetic programming (GP) algorithm. The proposed framework is called Wiki-ES (Wikipedia-based Evolutionary Semantics).

The traditional automated query learning systems usually represent both queries and documents using a bag-of-words approach. Moreover, the recent studies on IQBE paradigm have almost exclusively focused on finding the best evolutionary algorithms and fitness functions for learning boolean queries; see e.g. C\'ordon et al.~\cite{cordon03,cordon06}, Garc\'ia and Herrera~\cite{garcia08}, and L\'opez-Herrera et al.~\cite{lopez09a,lopez09b}. These developments have been well-motivated by the fact that the classical Boolean IR model is still broadly used, and there is considerable demand for query learning systems which can be run on top of any Boolean IR platform. However, restricting the query and document models to word-level information eliminates the possibility of leveraging human-level semantics on how the different {\it topics} and {\it concepts} are related. It should be noted that a query is composed of a number of concepts, and it represents the topic the user wants to search. To illustrate the difference between word based search and concept based search, consider a situation where a user is searching for information on a particular {\it topic}, for which he crafts a simple query ``economy AND espionage''. Then, suppose that a newly arrived document has {\it concepts} ``Trade secret'' and ``spying''. If we now ask a human reader to judge whether the document is about economic espionage, he would most likely find it relevant due to the close relationships between the concepts. However, if only word-level information is used, the boolean query will ignore the document as the original query words never appear.

In this paper, we focus on the benefits of using concepts instead of bag-of-words in query learning. As a test-bed for Wiki-ES system, we consider TREC-11 dataset with Reuters RCV1 corpus which provides a realistic example of a multi-domain news-stream. The experiments suggest that the concept-based approach is well-fitted to be used in conjunction with evolutionary algorithms. First of all, we observe that replacing tokens with Wikipedia's concepts yields significant improvement in filtering results as measured by precision and recall. The achieved improvements are considerable against other benchmarks as well, such as Support Vector Machines (SVM) and the decision-tree algorithm C4.5, which are well-known for their robust performance. Furthermore, when comparing the complexity of the query-trees produced by Wiki-ES against those of word-based IQBE-GP, we find that the use of concepts leads to simpler queries. 

The structure of this paper is following. Section~\ref{sec:contributions} summarizes the main contributions of the paper. Section~\ref{sec:preliminaries} gives a review on IQBE model for automated query learning, and how Wikipedia can be used as a source of semantic information. Section~\ref{sec:wikiESR} presents our framework Wikipedia-based Evolutionary Semantics (Wiki-ES). The co-evolutionary GP algorithm is presented in Section~\ref{sec:gpalgorithm}. Finally, Section~\ref{sec:experiment} summarizes the experimental results. 

\section{Contributions}\label{sec:contributions}

The key contributions of the paper are summarized in the following points:

\subsection{Use of Wikipedia semantics in query learning}

When a set of documents concerning a particular topic is to be retrieved from a database, it is common for a user to generate a query composed of token words. This query is used to decide the relevance of documents in a database by performing a search for the tokens in those documents. However, analyzing the problem from a user point of view, it is recognized that the user is not just interested in the documents containing the exact matching tokens, rather she is seeking all such documents which contain the concept represented by the token. This provides a motivation to work towards generating queries composed of concepts rather than tokens. Queries composed of concepts contain a wide human and society level knowledge, providing a better representation of the topic being searched. In this paper, we use Wikipedia semantics to convey the concept behind a token. There is no previous study to the knowledge of the authors, which utilizes the Wikipedia semantics to construct a concept based query. The efficacy of this transition from tokens to concepts, towards retrieval of documents, has been evaluated in the paper and its significance has been established.

\subsection{Development of a co-evolving GP}

Generating an accurate query for a search is often an iterative and tedious task to perform. However, if there is a set of documents available at hand, with each document marked  relevant or irrelevant, the task of query generation can be entirely avoided by directing the documents to an intelligent algorithm. Based on the relevance or irrelevance of the training documents, a concept based query can be learnt by the algorithm, saving the user from a monotonous task.
The paper contributes towards development of a co-evolving evolutionary algorithm specialized to generate concept based queries for document retrieval. The algorithm takes a set of training documents as input. Each document in the training set is marked as relevant or irrelevant by the user, based on which the algorithm produces concept based queries. The outcome of the algorithm is not a single query, rather a set of queries which are put together using a voting function. The use of multiple queries and a voting function leads to avoidance of any over-fit to the training set which may happen if only a single query is generated. Multiple queries produced by the algorithm, occupy different high fitness niches in the objective space and contribute towards the final decision for a document being relevant or irrelevant. Though genetic programming has been widely used for query construction, the implementations have usually involved, producing a single best fit token based query.

\subsection{Comparison of existing methodologies for Information Retrieval}

The paper performs a comprehensive comparison study of the existing methodologies. In addition to the proposed framework, other frameworks have also been evaluated on hundred different topics and the results have been presented. The concept based query construction approach has been implemented on the existing frameworks as well and significant improvement in results has been obtained for all the methodologies. Detailed evaluation results have been presented for the proposed framework against its closest competitor. In the extensive simulations performed, the proposed framework is found to outperform the other commonly used methodologies.

\section{Prelude: Wikipedia semantics and IQBE}\label{sec:preliminaries}

To provide an idea on the wealth of Wikipedia's semantic information and how that information can be utilized in query learning, we briefly discuss the recent innovations which leverage Wikipedia's link-structure to produce low-cost measures on the relationship between concepts and topics. In this section, we also summarize the recent developments in automated query learning. In particular, we consider the work inspired by evolution-based genetic algorithms, and the IQBE paradigm of Smith and Smith~\cite{smith97} and Chen et al.~\cite{chen98}.

\subsection{Wikipedia as a semantic knowledge-resource}

Research on ontology-based knowledge models has been largely motivated by their ability to provide unique definitions for concepts, their relationships and properties, which together create a unified description of a given domain. Having access to such structured information in machine-readable form has provided standardized ways for sharing common knowledge and, thus, enabled its efficient reuse in applications. Despite these advantages, the use of ontologies has been limited because of the large engineering costs that are unavoidable in manually built knowledge-resources. Furthermore, there is the difficulty of keeping the resources updated, in particular, when multiple domains are considered. As it is commonly known~\cite{sowa04,medelyan09}, even the most extensive ontologies, such as the Cyc ontology, have limited and patchy coverage. Therefore, the urgent need to find less expensive ways to describe concepts and their dependencies is well recognized. This has motivated research towards the use of socially or automatically constructed knowledge-resources.

When speaking of readily accessible multi-domain knowledge resources, the one that instantly comes into the mind is Wikipedia. Thanks to the activity of numerous volunteers, Wikipedia has rapidly matured into one of the largest repositories of manually maintained knowledge. Today, there are already over 3.3 million articles in English Wikipedia, and more arrive on a daily basis. The popularity of Wikipedia has also stimulated increasing research to investigate how the mountains of semantic information in Wikipedia can be harnessed for good uses; see Medelyan et al.~\cite{medelyan09} for a comprehensive review. As pioneering research in this field, we acknowledge the work done by Ponzetto and Strube~\cite{ponzetto06,ponzetto07,ponzetto07b,strube06}, Gabrilovich and Markovich~\cite{gabrilovich06,gabrilovich07,gabrilovich10}, Milne et al.~\cite{milne07,milne07b,milne08,milne09}, Medelyan et al.~\cite{medelyan08a,medelyan08b}, Nastase et al.~\cite{nastase08,nastase10}, and Mihalcea and Csomai~\cite{mihalcea07}, who have examined different ways of using Wikipedia to compute semantic relatedness between concepts and perform automated cross-referencing of documents.

\subsubsection{Wikipedia-concept}

In spite of the fact that Wikipedia does not really fulfill the criteria of being an ontology, a closer look at its structure reveals many similarities~\cite{hepp06}. By interpreting Wikipedia's articles as concepts, and by regarding the overall link-structure - including redirects, hyper-links, and category links - as relations, it is warranted to argue that Wikipedia is the largest semantic network available today. As nicely captured by Medelyan et al.~\cite{medelyan09}, Wikipedia provides a solid middle-ground between ontologies and classical thesauri ``by offering a rare mix of scale and structure''. Indeed, the recent developments suggest a number of ways in which Wikipedia can be used for extracting ontological knowledge; for example, see the Yago-ontology of Suchanek et al.~\cite{suchanek08} and WikiNet by Nastase et al.~\cite{nastase10}. 
\begin{figure}[h]
\centering
\includegraphics[scale=0.6]{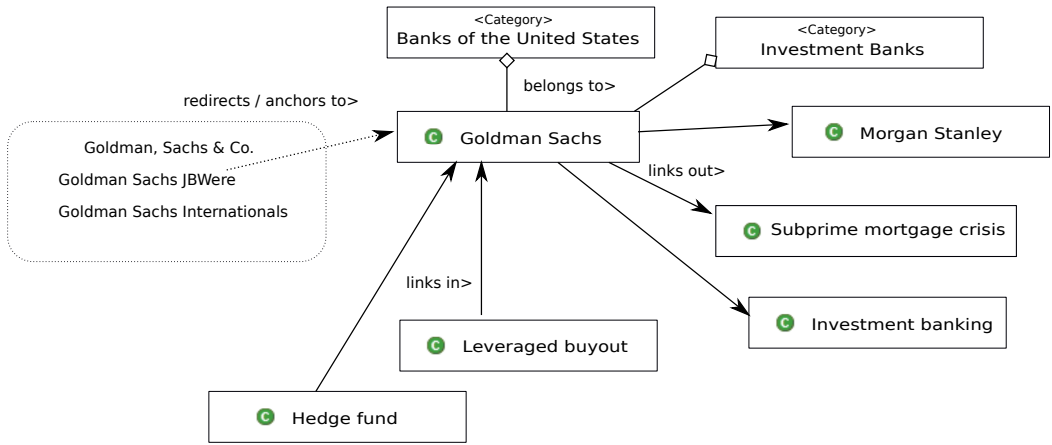} 
\caption{``Goldman Sachs'' as a Wikipedia-concept}\label{fig:concept}
\end{figure}

The primary feature that makes Wikipedia considerably richer in semantic knowledge than a conventional thesaurus is its dense internal link-structure. To illustrate the notion of Wikipedia-concept a bit more closely, let us consider, for instance, Wikipedia's article on ``Goldman Sachs'' (Figure~\ref{fig:concept}). Each Wikipedia-concept (article) belongs to at least one or more categories, which provide information about broader topics, hyponyms and holonyms. In this case, we find that Goldman Sachs belongs to categories such as ``Investment Banks'' and ``Banks of the United States''. Moreover, if the article's topic is sufficiently broad, then there also exists an equivalent category with the same title as the article. In addition to category-relationships, the articles have lots of hyper-links that represent semantic relationships between concepts. On average, each article refers to about 25 other articles. For instance, ``Goldman Sachs'' has links to many other banks (e.g. ``Morgan Stanley'') and financial concepts (e.g. ``Subprime mortgage crisis). These linkages can be exploited in various ways to mine knowledge on concepts and their relationships. Finally, to account for synonyms and alternative spellings of the article's name, each article has also a number of redirects that connect to the article. The redirects are complemented by anchors, which represent the words used within hyper-links that refer to the given article; and when several articles could be given the same name (e.g. Bank), then there is a disambiguation page that lists the alternative senses corresponding to that name. 

Therefore, considering the wealth of semantic information conveyed by Wikipedia, we find it natural to treat the Wikipedia-articles as equivalents for ontological concepts when modelling documents and queries. To formalise these ideas, we employ the following notation while referring to Wikipedia-concepts:
\begin{definition}[Wikipedia-concept]\label{def:concept}
Let $W$ denote the collection of Wikipedia-articles available for language $\Sigma$. Then a Wikipedia-concept is defined as an article $w\in W$, which is a uniquely identified representative of a certain concept. 
\end{definition}

Once we have the definition, there at least two questions that follow. The first one is concerned with concept-recognition. Clearly, it is not uncommon to find that several concepts may share the same textual representation. Thus, being able to resolve whether a certain concept is present in a document or not is a non-trivial problem. In the literature, this is commonly referred to as the wikification task~\cite{mihalcea07} or automatic topic-linking problem~\cite{milne08,medelyan08a}. This will be discussed more closely when outlining the content model used by Wiki-ES; see Section~\ref{sec:docmodel}. 

The second question, discussed in the following Section~\ref{sec:relatedness}, concerns the way semantic relatedness between any (concept,concept)-pair and (concept, document)-pair is measured. This needs to be resolved before we discuss the idea behind Wikipedia-based query rules and the way they are learned from example documents provided by a user. In particular, we need the notion of semantic relatedness while evaluating whether a document matches the given query or not.

\subsubsection{Measuring semantic relatedness}\label{sec:relatedness}

Although approaches to measuring conceptual relatedness based on corpora or WordNet have been around for quite long (McHale~\cite{mchale98} and Finkelstein et al.~\cite{finkelstein02}), the use of Wikipedia as a source of background knowledge is a relatively new idea. The first step in this direction was taken by Strube and Ponzetto~\cite{strube06}, who proposed their WikiRelate-technique that modified existing measures to better work with Wikipedia. This was soon followed by the paper of Gabrilovich and Markovitch~\cite{gabrilovich07}, who suggested explicit semantic analysis (ESA) to define a highly accurate similarity measure using the full text of all Wikipedia articles. 

The most recent proposal is, however, the Wikipedia Link-based Measure proposed by Milne et al.~\cite{milne07,milne08}, where only the internal link structure of Wikipedia is used to define relatedness. The approach is known to be computationally very cheap and has still achieved relatively high correlation with humans, which is why we have adopted it as a basis for the similarity measures used in this paper. The relatedness measure essentially corresponds to the Normalized Google Distance inspired by Cilibrasi and Vitanyi~\cite{cilibrasi07}
\begin{definition}[Link-relatedness (Milne et al.~\cite{milne07,milne08})]
Let $w_1$ and $w_2$ be an arbitrary pair of Wikipedia-concepts, and let $W_1, W_2 \subset W$ denote the sets of all articles that link to $w_1$ and $w_2$, respectively. The link structure -based concept-relatedness measure, $\text{link-rel}: W\times W \to [0,1]$ , is then given by
$$
\text{link-rel}(w_1,w_2)=\frac{\log{(\max{|W_1|,|W_2|})}-\log{(|W_1\cap W_2|)}}{\log{(|W|)}-\log{(\min{(|W_1|,|W_2|)})}}.
$$
\end{definition}

\begin{remark}
Although, this link-based relatedness measure is defined only for uniquely identified Wikipedia-concepts, it can be extended for calculating relatedness between any given pair of n-grams by using our knowledge about redirects and anchors attached to different concepts. 
\end{remark}

The underlying principle of $\text{link-rel}$ is rather simple: if two articles share a lot of same links, then they are likely to be highly related. For example, if we consider two major investment banks, such as ``Goldman Sachs'' and ``Morgan Stanley'', the {\it link-rel} yields a relatedness score of almost 80 percent due to the large number of financial concepts shared by both bank-articles. Whereas ``Goldman Sachs'' and ``Football'' are 0 percent related. Of course, these results are sensitive to the quality of the concept-articles' link-structure, and can thereby vary depending on the version of the Wikipedia being used. Nevertheless, when well-established articles are considered, and when speed is essential, we find that this kind of graph-based approach has proven to be a reasonably reliable way of measuring proximity between any arbitrary pair of concepts. 

So far, we have considered the computation of semantic relatedness in its conventional setup between two concepts. However, given our intention to use Wikipedia's relatedness information in matching queries and documents, it is perhaps more relevant to ask: how can we measure the relatedness between a document and a given concept? Or how likely is it for the given concept to appear in the document? For this purpose, we propose the following simple extension of the link-relatedness measure.
\begin{definition}[Document-concept relatedness]
Let $w\in W$ denote any Wikipedia-concept, and $d\in\mathcal{D}$ be an active document. The Wikipedia-based document-term -relatedness measure, $\text{d-rel}:W\times\mathcal{D}\to[0,1]$, is given by
$$
\text{d-rel}(w,d)=\max\{\text{link-rel}(w,\bar{w}) \ : \ \bar{w}\in \Lambda(d)\}
$$
where the document model, $\Lambda(d)$, is interpreted as the collection of Wikipedia-concepts detected in document $d$; see Section~\ref{sec:docmodel} for further discussion on document modelling. 
\end{definition}
Here, the use of maximum rather than sum-based operator such as average is a deliberate choice. Since {\it d-rel} is intended to be used in evaluating whether a document matches a given query, we do not want to allow any sum-operations to mask the presence of those concepts in a document which are not related to its central theme. To illustrate the idea, consider a single-concept query for documents on ``Industrial espionage''. Now, suppose that we receive a large document on car manufacturing, where most of the discussion is concerned with general economics and car models. However, the document still has a single paragraph on stolen trade secrets and car-prototype specifications. In order to prevent the document's main theme from hiding its relatedness to industrial spying, we choose to measure the relatedness by using the concept that is best associated with espionage. In this particular case, because trade secret is strongly linked to industrial espionage, it is natural to use their association to evaluate the overall relatedness between the document and the given query.

\subsection{Query learning problem}

The demand for automated query learning is driven by the difficulty of formulating effective queries that match the user's information needs. Finding appropriate search terms and conditions is generally hard even for expert users. Therefore, given a certain topic, the task of query learning systems is to help the user to find a query definition with improved precision and recall. As the size of world's information base is growing at a staggering rate, the problem is becoming increasingly pressing. To alleviate it, a large number of competing solutions for query formulation have been proposed in response. As suggested by C\'ordon et al.~\cite{cordon03}, these can roughly be categorized into three baskets: (1) term learning; (2) weight learning; and (3) query-structure learning. 

The commonality of the approaches is their reliance on some form of relevance feedback, where the system elicits (possibly iteratively) a set of feedback statements from the user. In the first two model categories, relevance feedback is used for modifying the user's previous query by removing or adding terms and adjusting their weights to better reflect the user's relevance judgements. For example, many of the probabilistic models and document-vector modification models belong to these categories; see e.g. Salton and Buckley~\cite{salton90} and Rocchio~\cite{rocchio71}, Yang and Korfhage~\cite{yang94}, Horng and Yeh~\cite{horng00}, and Boughanem et al.~\cite{boughanem99,boughanem02}. 

Our focus is on the third category, query-structure learning, which takes the learning process one step further in the context of boolean or fuzzy boolean queries.  It not only attempts to infer the terms that are most appropriate for representing a given query but also tries to learn the query's structure, i.e. it determines how the boolean operators AND $(\wedge)$, OR $(\vee)$, and NOT $(\neg)$ should be used to join the different concepts. In many texts, {\it query learning} is considered as a reserved word for representing this third type of query definition, where both the functional form and concepts of the query are free variables; see e.g. C\'ordon et al.~\cite{cordon03,cordon06}, L\'opez-Herrera et al.~\cite{lopez09a,lopez09b}  and their references. The IQBE paradigm (Section~\ref{sec:iqbe}) and the Wiki-ES system introduced in this paper are mainly viewed as structural query learning models. Therefore, for the rest of this paper, we will use the following general definition to refer to query learning problem.
\begin{definition}[Query learning problem]\label{def:query}
Let $C$ be a set of admissible concepts, and let $Q$ denote the space of all admissible queries which can be formed using concepts in $C$. The query learning task is to find that boolean expression from the set $Q$ which best represents the user's information needs by applying the following syntactic rules:
\begin{enumerate}
\item Atomic query (single concept): $\forall \ q=c_i\in C \to q\in Q$ 
\item Composition using AND: $\forall \ q,p\in Q \to q\wedge p\in Q$
\item Composition using OR: $\forall \ q,p\in Q \to q\vee p\in Q$
\item Negation: $\forall \ q\in Q \to \neg q\in Q$
\end{enumerate} 
The space of admissible queries $Q$ consists of all the queries obtained by applying the above set of rules.
\end{definition}
There are many ways to approach the above problem - both with and without the use of semantic knowledge. At this stage, we notice that the definition remains deliberately abstract by not specifying how the set of concepts should be understood and how the learnt queries be matched against documents. Of course, when classical boolean queries using the bag-of-words approach are considered, the answer is quite straightforward. However, when the atoms of a query are uniquely defined concepts, it is no longer self-evident how the query should be evaluated. In fact, as we find out in Wiki-ES model, the performance differences between concept-based and word-based approaches follow from the way concept-relationship information is used while matching documents with learnt queries. 

\subsection{IQBE - Inductive Query By Example}\label{sec:iqbe}

\begin{figure}[!ht]
\centering
\includegraphics[scale=0.5]{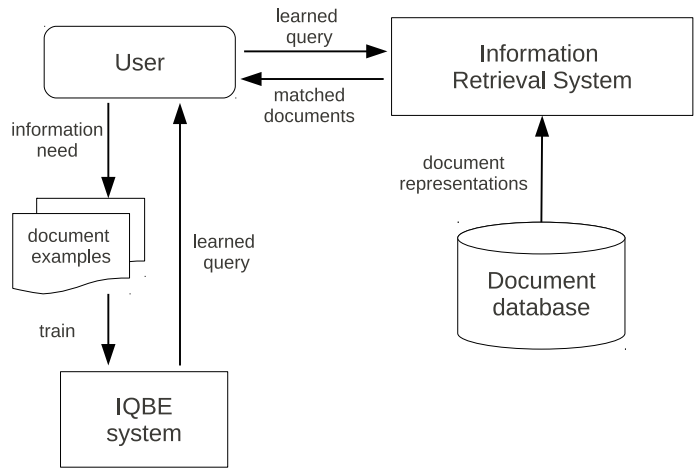} 
\caption{A general IQBE architecture}\label{fig:iqbe}
\end{figure}

One of the best known bag-of-words based methods for solving the query learning problem~\ref{def:query} is the Inductive Query By Example (IQBE) framework originated by Smith and Smith~\cite{smith97} and Chen et al.~\cite{chen98}. The idea behind IQBE paradigm is in principle very similar to relevance feedback; both of them require explicit relevance statements from the user to guide the retrieval process. In IQBE, the user provides the system with a collection of sample documents (positive/negative examples) from which an algorithm learns the terms and the boolean operators joining them, such that the obtained query best represent the user's information need. However, instead of modifying an existing query iteratively, the system performs only a single run to generate a fresh query from the scratch. Once the learnt query is available, it can be executed on any boolean information retrieval system (IRS). Such portability of queries can be considered as one of the characteristics that distinguishes IQBE systems from general relevance feedback. In descriptions of IQBE architecture, this is commonly emphasized by presenting IQBE system as a separate unit outside the IRS; see L\'opez-Herrera et al.~\cite{lopez09a} and Figure~\ref{fig:iqbe} for descriptions of a general IQBE system.

In IQBE framework, the query learning task is viewed as a large optimization problem, where the search space consists of all possible queries that can be presented to the IRS. Therefore, recognizing the high dimensionality of this problem, it is no surprise that the IQBE approaches usually rely on some form of evolutionary computation. In particular, following the early studies by Kraft et al.~\cite{kraft95} and Smith and Smith~\cite{smith97}, genetic programming~\cite{koza92} has gained ground as a robust choice for query learning. Recently,  a number of frameworks based on multiobjective genetic programming have also been examined. Due to the fact that the performance of an IRS is mostly evaluated in terms of precision and recall, it appears natural to consider query learning as an inherently multiobjective problem. For interesting applications of multiobjective evolutionary algorithms, see e.g. C\'ordon et al.~\cite{cordon06} and L\'opez-Herrera et al.~\cite{lopez09a,lopez09b}. 

As discussed by Tamine et al.~\cite{tamine03}, the popularity of evolutionary algorithms is largely explained by their implicit parallelism which allows them to search different regions of the solution space simultaneously. It is also argued that evolutionary algorithms are less sensitive to the quality of the initial query. Whereas classical relevance feedback methods, such as Rocchio~\cite{rocchio71}, perform poorly if the initial query fails to retrieve relevant documents. The probabilistic exploration induced by evolutionary algorithms permits them to search unexplored areas independent of the initial query~\cite{cecchini08}. Hence, the use of evolutionary algorithms is a well-justified choice for query learning as non-expert users can rarely find a good query on a first try when more complicated topics are considered.

Although the automated query learning problem has stimulated a lot of interest over the past few years, it is noteworthy that majority of the development has concentrated on improving learning algorithms rather than examining the role of query and document representations. However, recognizing the fact that the use of semantic information has transformed many natural language processing applications~\cite{zhuge10}, we consider it worthwhile to work towards the development of a Wikipedia-concept based approach which would enhance automated query learning.

\section{Wiki-ES: Learning semantic queries with Wikipedia}\label{sec:wikiESR}

In this section, we present the Wiki-ES (Wikipedia-based Evolutionary Semantics) framework for automated query learning. The approach is based on the Genetic Programming (GP) paradigm, which is a potent tool in artificial intelligence for performing program induction. In GP, the idea is to use the principles of evolutionary computation to intelligently search the space of possible computer programs for finding an individual that is highly fit for solving the problem at hand. In effect, one could say that the purpose is to get the machine to generate a solution to the problem without being explicitly programmed~\cite{koza92}. For example, in our case we want the Wiki-ES system to learn a program (i.e. query) that leads to recovery of a high number of relevant documents while keeping the irrelevant documents aloof. The learning process is driven by the evolutionary pressure which guarantees that only the fittest individuals among all potential query candidates survive.

\subsection{Wiki-ES framework overview}\label{sec:framework}

A bird-eye's view of the Wiki-ES framework resembles the architecture of the IQBE paradigm (see Figure~\ref{fig:iqbe}), where the idea is that the system is able to learn an optimal query by using just a small set of sample documents  that represent the user's current topic or information need. On the surface, this sounds simple. However, when examining the steps involved in the learning process, it becomes clear that a number of choices, ranging from the choice of query and document models to the choice of the genetic procedure, have large impacts on the outcome. 

\begin{figure}[!ht]
\centering
\includegraphics[scale=0.5]{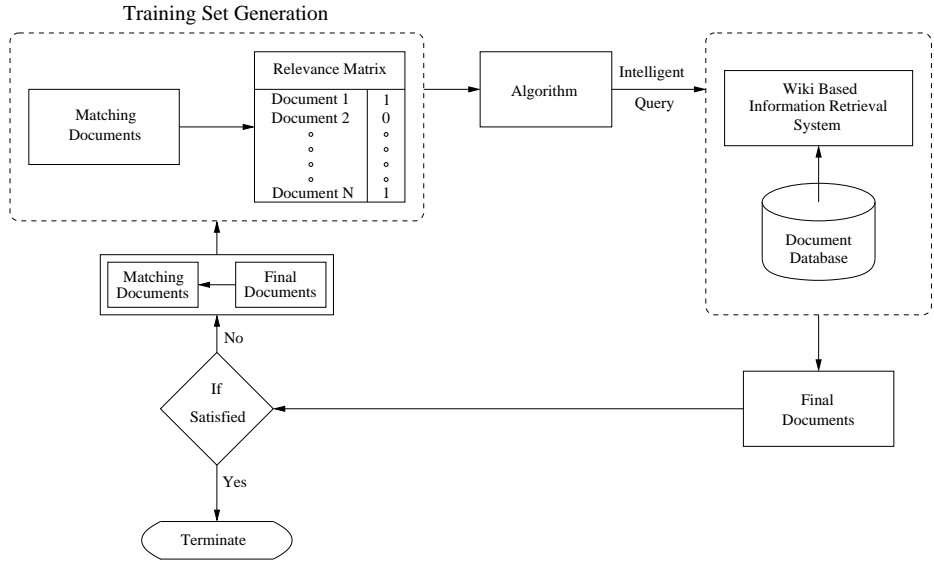} 
\caption{Wiki-ES flowchart}\label{fig:wikiESRflowchart}
\end{figure}

To illustrate the way Wiki-ES approaches the query learning problem, let us consider an example where a user seeks to define a query that picks up all the documents on economic espionage but ignores the ones on politics or military espionage. Then, we can split the Wiki-ES process into the following steps (see Figure~\ref{fig:wikiESRflowchart}):
\begin{enumerate}
\item {\it Training data generation}: Suppose that the user has already found a bunch of documents that she considers highly relevant for the topic and also a collection of documents that are concerned with espionage but are more about military spying than industrial espionage. Then, the training data set is defined as a relevance matrix, where each sample document is given a boolean value to represent its relevance for the topic (1=relevant, 0=irrelevant).
\item {\it Learning an intelligent query}, i.e. the Wiki-ES rule: In the learning step, the training data set is given to the GP-algorithm to find an optimal Wiki-ES rule to describe the topic. Each Wiki-ES rule consists of a number of queries, which allows the rule to take into account not only the concepts which appear directly in the query-expressions but also the ones which are strongly related to them. A detailed description of the rules is given in Section~\ref{sec:querymodel}. The GP-algorithm is described in Section~\ref{sec:gpalgorithm}.
\item {\it Feeding the Wiki-ES rule and documents to the Wiki-based Information Retrieval System (WIRS)}:
Once the optimal Wiki-ES rule is known, it can be given to a matching subsystem which evaluates the query against the incoming documents. In Wiki-ES framework, this task is handled by WIRS module, which consists of two subsystems: the document modeling subsystem and the rule-matching subsystem.
\begin{enumerate}
\item {\it Document modeling subsystem}: Before the incoming documents can be matched against the Wiki-ES rules, they are passed through a wikifier and a named-entity recognizer (NER). The resulting profile, expressed in terms of the identified Wikipedia concepts and named-entities, can then be used to represent the document contents when matching against Wiki-ES rules; see Section~\ref{sec:docmodel} for description of the document model.

\item {\it Rule-matching subsystem}: The rule-evaluator in WIRS module provides a matching subsystem for deciding whether a given document matches the currently active semantic rule or not. In Wiki-ES framework, it is hence the responsibility of the rule-evaluator to utilise Wikipedia's concept-relatedness information while determining whether the query concepts are present in the active document - either directly or indirectly. The way how the rule-evaluator operates is described in Section~\ref{sec:querymodel}.

\end{enumerate}

\item {\it Returning the filtered documents to user}: The documents, that are found to match the active Wiki-ES rule, are returned to the user. If the user is satisfied with the retrieved documents, then the process terminates. Otherwise, a new training data set is created using the final documents and the initial matching documents, and the system returns to step 1.

\end{enumerate}

Having provided a rough schematic overview of the Wiki-ES model, we are now ready to explain more closely the underpinnings of the Wiki-based Information Retrieval System (WIRS). The remainder of this section is organized as follows. First, in Section~\ref{sec:docmodel}, we discuss the Wikipedia-based content model used within the document modeling subsystem of WIRS. Then, the Section~\ref{sec:querymodel} continues by outlining the structure of the rule-matching subsystem in WIRS. In particular, we define what Wiki-ES rules are and how they are evaluated. Finally, we formalise the Wiki-ES learning problem in Section~\ref{sec:optim}. The details of the learning algorithm are treated separately in Section~\ref{sec:gpalgorithm}.

\subsection{Wikipedia-based document model}\label{sec:docmodel}

In Wiki-ES framework each document is represented by a collection of Wikipedia-concepts that are identified from its contents. The approach builds on the wikification technique proposed by Milne et al.~\cite{milne08} and Medelyan et al.~\cite{medelyan08a}, where a two-stage classifier is utilised to recognize those terms in the document which should act as Wikipedia-concepts. However, the model employed here extends the wikification-process by splitting the found concepts into two categories, general Wikipedia-concepts and named-entity concepts, using a named-entity recognizer. 

To explain the rationale for this modification, consider, for example, a named entity ``Goldman Sachs'' and a general concept ``Investment banking''. Now, to say that a certain document discusses Goldman Sachs requires that the bank's name is explicitly mentioned. On the other hand, if we say that a document is about investment banking, it is sufficient to find a collection of investment banking related concepts rather than the exact concept name to identify the document as relevant. Clearly, the different nature of general concepts and named-entities should be taken into account when specifying the sensitivity of the Wiki-ES model to different concept types. Hence, in Wiki-ES, each document is interpreted as a pair of two collections: the named-entities and other Wikipedia concepts.
\begin{definition}[Wiki-ES document model]
Let $D$ be the space of documents, and $W$ denote the collection of Wikipedia-concepts. The document model is defined as the mapping 
$$
\Lambda: d\in D \mapsto (N_d,G_d)\subset W\times W,
$$
where $N_d$ and $G_d$ denote the sets of named-entities and general Wikipedia-concepts found in the document $d$. 
\end{definition}
So, for example, if a document $d\in D$ contains Wikipedia-concepts, $\{$ Investment banking, Goldman Sachs, Morgan Stanley, Mortgage, Credit $\}$, we simply present the document model as split into two parts, $\Lambda(d)=(N_d,G_d)$, where $N_d=\{$Goldman Sachs, Morgan Stanley$\}$, $G_d=\{$Investment banking, Mortgage, Credit$\}$. 

The document model $\Lambda$ is implemented in two stages: wikification and named-entity recognition. Once the usual preprocessing steps have been carried out, the first stage is to identify all Wikipedia-concepts present in the document. At this stage, no separation between named-entities and general concepts is made. Here, identification is done using the wikification (or cross-referencing) technique proposed by Milne et al.~\cite{milne08}, where a sequence of two classifiers is run to detect which terms should be linked to Wikipedia and to which Wikipedia-concepts do they correspond. 

The second step, named-entity recognition (NER), is done by using the Conditional Random Fields (CRF)-based classifier proposed by Finkel et al.~\cite{finkel05}. An advantage of this model is that the system is able to augment non-local information which allows construction of long-distance dependency models and enforcement of label consistency. In this approach, the set of Wikipedia-named-entities is identified by examining the overlap between the terms that have been picked-up by the wikification step and those recognized by the NER-classifier.  

\subsection{Query model: structure and matching of Wiki-ES rules}\label{sec:querymodel}

As mentioned in Section~\ref{sec:framework}, each Wiki-ES rule can be viewed as a composition of a number of queries. The Wiki-ES rule has an underlying structure that is essentially different from what is seen in ordinary boolean queries. To provide a more accurate picture, we formalise the definition of Wiki-ES rule as a voting system where several concept based queries go for a voting and the weighted sum of their votes is taken to represent the relevance of a document. 

The presentation of the Wiki-ES model is structured as follows. First, we define the Wiki-queries that are used as building blocks in Wiki-ES rule (Section~\ref{sec:wikiquery}). Thereafter, in Section~\ref{sec:wikiESRrule}, we introduce a fitness-measure for evaluating the quality of individual queries, and discuss how a voting system can be used to combine the output of several Wiki-queries to generate a Wiki-ES rule. Section~\ref{sec:optim} summarizes the Wiki-ES learning problem. We also discuss the benefits of constructing the Wiki-ES rule as a voting system instead of using the individual queries directly. 

\subsubsection{Building blocks of Wiki-ES rules}\label{sec:wikiquery}

Now, we begin by outlining the types of boolean queries used as building blocks  for the Wiki-ES rule. To distinguish these from ordinary term-based queries, we refer to them as Wiki-queries (concept-based queries) hereafter. Unlike an ordinary boolean query, a Wiki-query consists of two parts. In addition to the query-expression, each Wiki-query also contains a specialized evaluator function which allows the query to utilize Wikipedia's concept-relatedness information when it is matched against documents. 

\begin{definition}[Wiki-query]
A Wiki-query $q: D\to \{0,1\}$ is defined by a pair $(e,\delta)$, where
\begin{itemize}
\item[(i)] the first component, $e$, is an ordinary query-expression that is defined in terms of Wikipedia-concepts $V\subset W$ and the standard boolean operators by following the syntactic rules outlined in~\ref{def:query}; and
\item[(ii)] the second component, $\delta:V \times D \to \{0,1\}$, is a concept-evaluator function given by~\ref{def:conceptevaluator}, which determines whether a concept $v\in V$ is present in any given document $d\in D$. 
\end{itemize}
When matching the given query $q=(e,\delta)$ against any document $d$, the value of the query $q(d)$ is obtained by replacing each concept $v\in V$ in the query expression $e$ with the corresponding value $\delta(v,d)$ given by the concept-evaluator. 
\end{definition}
\begin{example}
Let $q$ be defined by $(e,\delta)$. If $e=v_1r_1v_2r_2\cdots r_{k-1}v_k$, and $v_i\in W$, $r_i\in\{\wedge,\vee,\neg\}$ for all $i=1,\dots,k$, then the value of the query amounts to $q(d)=\delta(v_1,d)r_1\delta(v_2,d)r_2\cdots r_{k-1}\delta(v_k,d).$
\end{example}
\begin{definition}[Concept-evaluator]\label{def:conceptevaluator}
The concept-evaluator function, $\delta:V\times D\to \{0,1\}$, whose purpose is to account for Wikipedia's concept-relatedness information when evaluating Wiki-queries, is given by
\[
\delta(v,d)=
\begin{cases}
1 & \text{if $v\in \Lambda(d)$},\\
1 & \text{if $v\in \text{Rel}(d)$}, \\
0 & \text{otherwise}
\end{cases}
\]
where 
$$
\text{Rel}(d)=\{v\in V : \text{d-rel}(v,d)>c_{\text{rel}}(v)\},
$$ 
and $c_{\text{rel}}>0$ is a threshold function controlling the acceptance sensitivity by relatedness criteria. The threshold for document-concept relatedness function (d-rel) depends on the type of concept, i.e. whether it is a named-entity or general Wikipedia-article. If $\Lambda(d)=(N_d,G_d)$, we have
$$
c_{\text{rel}}(v)=
\begin{cases}
c_1 & \text{if $v\in N_d$}, \\
c_2 & \text{if $v\in G_d$}. 
\end{cases}
$$
Each sensitivity threshold is chosen based on training data. The purpose of the distinction between named-entities and general concepts is to allow stricter thresholds for named-entities which have narrower definitions than general concepts.
\end{definition}

To illustrate the underlying idea, consider a simple Wiki-query, $q=(e,\delta)$, where the query-expression 
$$
e=  Lawsuit \wedge (Espionage\vee TradeSecret) \wedge BMW 
$$ 
requests for documents on industrial espionage that are concerned with BMW. Now, suppose that the following document $d$ is received:
\begin{quote}
A civil court in Hamburg will give its verdict on Tuesday on a hearing called by Spiegel, a leading German magazine. Spiegel is trying to lift an injunction from VW preventing it from repeating allegations of corporate spying against Mr Lopez...The documents include top-secret details of Opel's new small car project, coded the O-car, which is to rival Volkswagen's planned Chico.
\end{quote}
Once the document has been profiled, it can be evaluated against the query expression. In this case, during the concept-evaluation step, we find that $\delta(Lawsuit,d)=1$ because the terms ``civil court'' and ``allegation'' point to \textit{Lawsuit}, and similarly we have $\delta(Espionage,d)=1$ because ``spying'' is a redirect to \textit{Espionage}. However, the evaluation of the concept \textit{TradeSecret} and the named-entity concept \textit{BMW} turn out to be more problematic as they will depend on the acceptance-sensitivity function ($c_{\text{rel}}$). 

Let us first consider the \textit{TradeSecret}-concept. To determine whether \textit{TradeSecret} is present in the document, we need to examine its relatedness to other concepts that have been identified from the document. In the above excerpt ``top-secret'' is recognized as \textit{ClassifiedInformation} which is strongly related to \textit{TradeSecret}, therefore the decision boils down to the comparison of these two concepts. Here, $\delta(TradeSecret,d)$ equals 1 only if the acceptance sensitivity $c_{\text{rel}}(TradeSecret)$ is less than the link-relatedness measure between \textit{TradeSecret} and \textit{ClassifiedInformation}.

So far, it seems that the given excerpt is almost a match provided that the last concept, \textit{BMW}, is also recognized as related to the document. Now, the acceptance sensitivity parameter for named-entities $c_1$ is set at a reasonably strict-level, say 0.95, to ensure that named-entities are not as broadly defined as the general concepts. For example, one would observe a very high relatedness between \textit{BMW} and \textit{VW} as they are both German car manufacturers with almost similar link-structures. However, mixing these two would be a serious error from the user's point of view. Therefore, being able to define acceptance sensitivities separately for named-entities and general Wikipedia-concepts proves to be a useful tool. Eventually, due to high value of $c_1$, we deduce that $\delta(BMW,d)=0$, and therefore the document is considered to be irrelevant. 

\subsubsection{Wiki-ES rule}\label{sec:wikiESRrule}

Having introduced Wiki-queries, we are now ready to explain how they are combined to generate a Wiki-ES rule. For this purpose, we define two additional functions: (i) a fitness-function for measuring the quality of individual Wiki-queries; and (ii) a voting function for summarizing the output of a group of Wiki-queries into a single measure.

\begin{definition}[Fitness of Wiki-query]\label{def:fitness}
Let $Q$ denote the space of admissible Wiki-queries. The fitness-function for a Wiki-query $q\in Q$ is defined as the mapping, $F:(q,D_t)\mapsto c\in[0,1]$, which corresponds to the F-score within a given set of evaluation documents $D_t\subset D$:
$$
F(q,D_t)=\frac{2 P(q,D_t)R(q,D_t)}{P(q,D_t)+R(q,D_t)},
$$
where $P(q,D_t)$ is the precision of the query in the document set $D_t$, and $R(q,D_t)$ is the recall of the query, respectively. By denoting the relevance of a document $d\in D_t$ by $r(d)\in\{0,1\}$, precision and recall are defined as 
\[
P(q,D_t)=\frac{\sum_{d\in D_t}r(d)q(d)}{\sum_{d\in D}q(d)} \quad\text{and}\quad R(q,D_t)=\frac{\sum_{d\in D_t}r(d)q(d)}{\sum_{d\in D}r(d)}.
\]
\end{definition}

Now, suppose that instead of having a single query to describe the user's information need, we have several complementary queries for the same topic, where each query represents a part of the user's need. In order to benefit from the diversity provided by the multiple query representation, we first need to resolve how the potentially conflicting results from different queries can be combined into a single document-relevance measure. Given the above F-score as a fitness-measure for evaluating the quality of each individual Wiki-query, a natural approach for dealing with this ``query fusion'' problem is to consider the following voting function where each query contributes to the overall relevance judgement according to its relative fitness:

\begin{definition}[Voting function]
Let $A\subset Q$ be a finite collection of Wiki-queries. A voting function $\mu_A:D\to [0,1]$ is given by
\[
\mu_A(d)=\frac{\sum_{i=1}^{|A|}F_iq_i(d)}{\sum_{i=1}^{|A|}F_i},
\]
where $F_i=F(q_i,D_t)$ is the fitness of query $q_i$ evaluated with respect to a training document set $D_t\subset D$.
\end{definition}
\begin{remark}
The voting function $\mu_A$ can be also used for ranking the documents based on their relevance to the given topic. However, the use of rank-order information is left as a direction for further research.
\end{remark}

The value of the voting function has an interpretation as the joint-relevance of a document, where the judgement is based on several alternative queries that describe the given topic. If the value of the voting function is greater than 0.5, then the document is considered relevant, otherwise it is considered irrelevant. Using this weighted contribution, the information from several queries is taken into account, which helps to reduce the risk of overfitting the training document set with a single query. This discussion is formalised by the following definition of the Wiki-ES rule.

\begin{definition}[Wiki-ES rule]\label{def:wikiESR}
Let $Q$ denote the space of admissible boolean queries formed using Wikipedia-concepts, and let $\mu_A$ be a voting function that evaluates the document-relevance based on a finite set of Wiki-queries, $A\subset Q$. Now, the Wiki-ES rule  is defined as the function $\bar{q}_A:D\to\{0,1\}$:
\[
\bar{q}_A(d)=
\begin{cases}
1 & \text{if $\mu_A(d)>0.5$},\\
0 & \text{otherwise}
\end{cases}
\] 
and the space of admissible Wiki-ES rules is given by $\bar{Q}=\{\bar{q}_A \ | \ A\subset Q \}$, where $A$ denotes any finite set of Wiki-queries.
\end{definition}
\begin{remark}
At this point, it is worthwhile to note that any Wiki-query can be viewed as a Wiki-ES rule, i.e. $Q\subset \bar{Q}$, because for every Wiki-query $q_0\in Q$, we have $\bar{q}_{\{q_0\}}\in \bar{Q}$. Hence, the Wiki-ES rules provide a natural extension of the Wiki-queries. 
\end{remark}

\subsection{Wiki-ES as an optimization problem}\label{sec:optim}

As discussed in Section~\ref{sec:iqbe}, the query learning task can be viewed as a large optimization problem, where the search space consists of all possible queries that can be presented to the IRS.  However, instead of considering optimization over the space of admissible Wiki-queries, we convert the query learning task into the problem of finding an optimal Wiki-ES rule which maximizes F-score with respect to the given collection of training documents.

\begin{definition}[Wiki-ES learning problem]
Let $D_t\subset D$ be the set of training documents for which user has given relevance statements, and let $\bar{Q}$ denote the space of Wiki-ES rules. The learning problem is given by
$$
\bar{q}^{\star}=\argmax_{\bar{q}\in \bar{Q}}F(\bar{q},D_t)
$$
where $F:(\bar{q},D_t)\mapsto c\in[0,1]$ is the Wiki-ES fitness function, which corresponds to the F-score within the training document set $D_t$; see Definition~\ref{def:fitness}.
\end{definition}

The rationale for defining the learning problem in terms of Wiki-ES rules instead of Wiki-queries stems from the following reasons. The first one is the multimodality of the user's relevance function. As pointed out by Tamine et al.~\cite{tamine03}, the relevant documents corresponding to the same topic can be dispersed into different regions of the document space, and thereby have quite different profiles. This implies that in order to recover the relevant documents it is necessary to explore the document space in a number of directions at the same time. Therefore, given the definition of a Wiki-ES rule as a voting system, it appears to be a natural solution for the multimodality problem as it utilises a number of Wiki-queries while making the retrieval decisions. 

The use of Wiki-ES is also motivated by the fact that unlike classical methods, GP-based approaches always operate with a population of queries rather than a single query. Therefore, we are likely to obtain better results by using several individuals from the population to represent the solution rather than rely on a single query candidate. Hence in order to solve the above optimization problem, we have chosen to use a co-evolutionary GP approach, where multiple subpopulations are evolved simultaneously to produce Wiki-queries that can be combined to produce an optimal Wiki-ES rule. The details of the algorithm are provided in Section~\ref{sec:gpalgorithm}.

\section{Wiki-ES GP-algorithm}\label{sec:gpalgorithm}

The aim of the proposed GP algorithm is to generate better fit queries using a mechanism inspired by biological evolution~\cite{poli08}. The approach is population based, where each individual represents a Wiki-query. The idea behind the technique is that, for a given population of individuals, the environmental pressure causes natural selection leading to a rise in the fitness of the population. Once the genetic representation of a query and the fitness function is defined, the algorithm proceeds to initialise a population of queries randomly. The population of Wiki-queries is then improved through repetitive application of Selection, Crossover, Mutation and Replacement. To ensure sufficient diversity and reduce the risk of over-fitting the training set, the population is evolved in a number of co-evolving sub-populations. The Wiki-ES rules are then formed by collecting the fittest individuals from each sub-population to form the set of queries that participate in the voting function.  

The remainder of this section is structured as follows. First, Section~\ref{sec:geneticrep}, begins by providing the genetic representation of Wiki-queries as syntax trees. Next, the initialization of the query populations is discussed in Section~\ref{sec:initialization}. Fitness assignment and the production of new queries is covered in Sections~\ref{sec:assign} and~\ref{sec:produce}. Finally, the structure of the evolutionary algorithm is presented in Section~\ref{sec:algodesc}, which is followed by a short discussion on the formation of Wiki-ES rules in Section~\ref{sec:formation}.

\subsection{Genetic Representation}\label{sec:geneticrep}
\begin{figure}[hbt]
\begin{center}
\epsfig{file=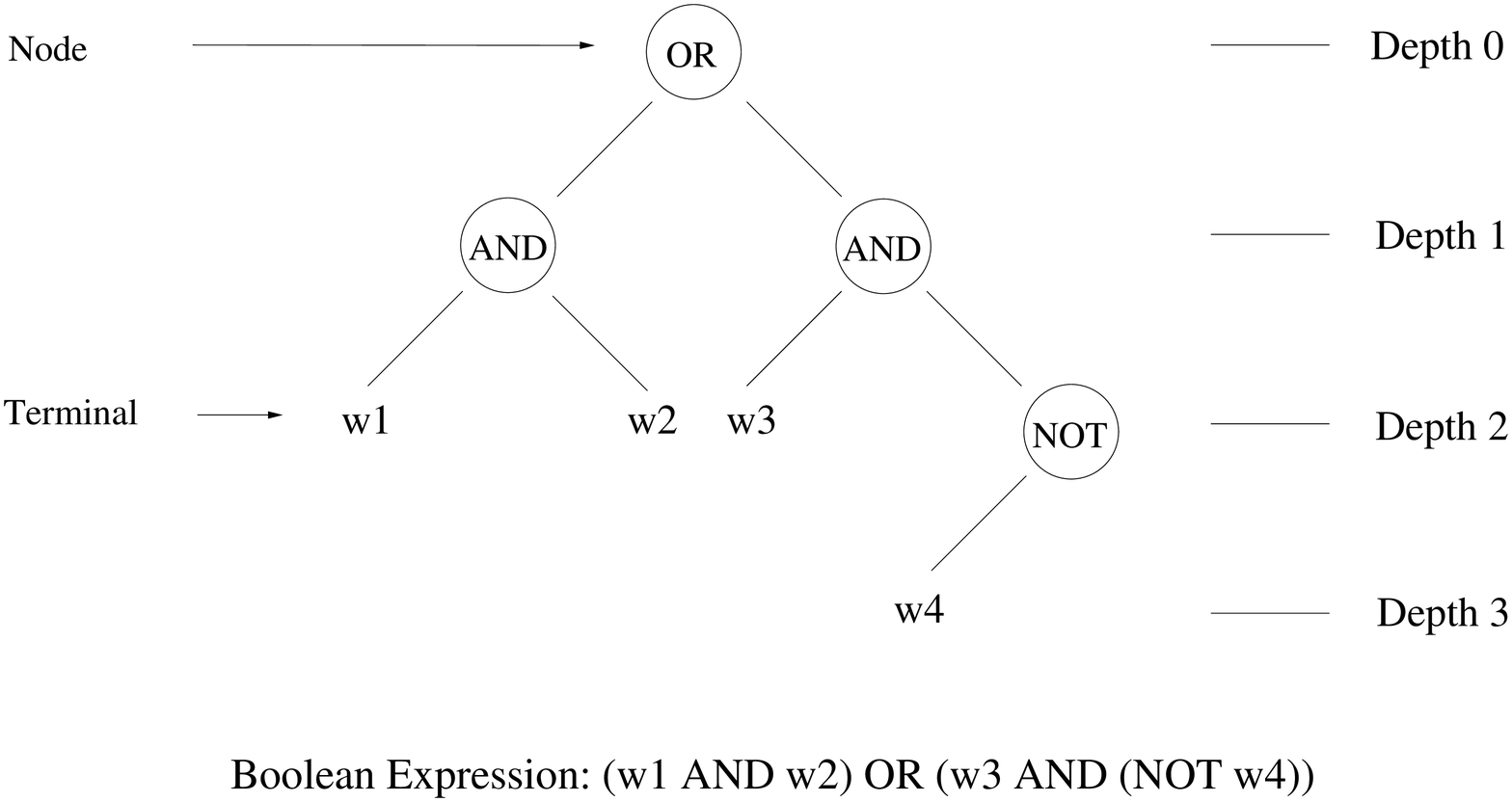,width=0.75\linewidth} 
\caption{Genetic Representation}
\label{fig:representation}
\end{center}
\end{figure}

Each query is expressed as a syntax tree with the nodes acting as boolean operators and the the terminals as the concepts; see Table~\ref{tab:gpcomponents} for correspondence between the common GP components and the Wiki-queries. Figure~\ref{fig:representation} shows one such query which acts as an individual in the population. The query shown in the figure is composed of four concepts, $\{w_1, w_2, w_3, w_4\}$, and the basic boolean operators, $\{AND, OR, NOT\}$. The tree represents a boolean expression $(w_1 \wedge w_2) \vee (w_3 \wedge (\neg w_4))$. Such a query will lead to the selection of those documents from the library which either contain the concepts $w_1$ and $w_2$ or it contains the concept $w_3$ but not $w_4$. Each tree has a depth which is a representative of the size of a tree. The depth of a tree is the number of branches traversed to reach the deepest terminal. The tree in the Figure~\ref{fig:representation} has $w_4$ as the deepest terminal and the depth of the tree is $3$. It should be noted that the depth of a root node is 0.

\begin{table}[h]
\caption{The interpretation of GP-components in Wiki-query context}\label{tab:gpcomponents}
\centering
\begin{scriptsize}
\begin{tabular}{| l | l |} 
\hline
GP component & Meaning in Wiki-query \\
\hline
Terminals (leaf nodes) & Wikipedia-concepts in a query-tree \\
Functions (non-leaf nodes) & Boolean query operators (AND, OR, NOT) \\
Fitness function & The objective function (F-score) in the  \\
 & query learning problem  \\
Reproduction, crossover, and & Genetic operators for driving the development \\
mutation & of Wiki-queries according to the evolutionary  \\
 & principles. \\
\hline
\end{tabular}
\end{scriptsize}
\end{table}

\subsection{Population Initialization}\label{sec:initialization}
\begin{figure}[hbt]
\begin{center}
\epsfig{file=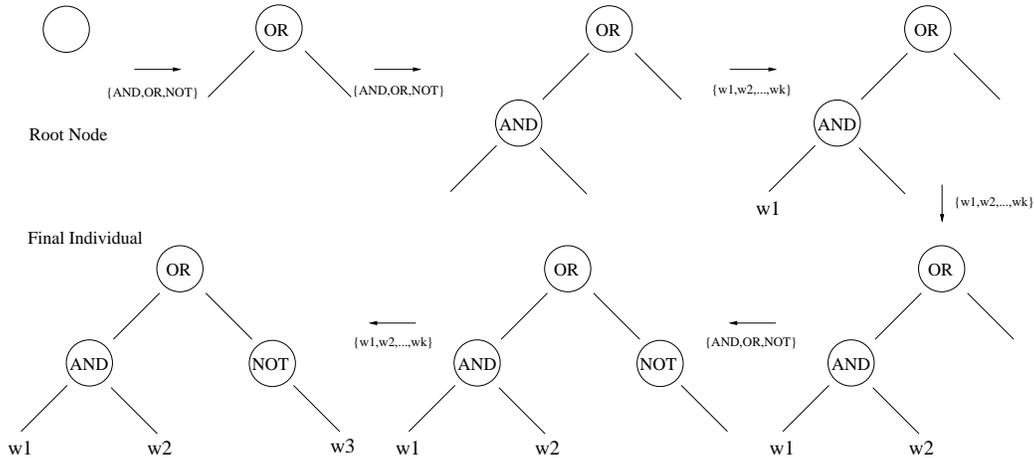,width=1.00\linewidth} 
\caption{Random initialisation of an individual}
\label{fig:initialise}
\end{center}
\end{figure}

Like in any evolutionary algorithm, the initial population individuals are generated randomly in genetic programming. The maximum depth ($d_{max}$), an individual can have, is given as input. A number $d$ is chosen randomly from the set $\{1,2,3 \ldots, d_{max}\}$. The chosen number becomes the depth of the tree (individual) to be initialized. Starting from the root node, an operator is chosen randomly from the set $O = \{AND, OR, NOT\}$, and placed at the node. If the node turns out to be $AND$ or $OR$, then two sub nodes are created; otherwise a single sub node is created. The procedure is repeated for each of the sub nodes and the tree size grows. At a depth $d-1$, a terminal should be chosen to terminate the growth of the tree. Therefore, random choice is made from the set $W_0 = \{w_1, w_2, \ldots, w_k\}$ and the concept is placed at the terminal. This completes the procedure to generate a single individual. Following a similar procedure, a number of individuals equal to the population size $N$ are generated; the next step is to assign fitness to each individual. Figure~\ref{fig:initialise} shows the steps involved in initialising an individual of depth $2$.

\subsection{Fitness Assignment}\label{sec:assign}
As already mentioned, the set $W_0 = \{w_1, w_2, \ldots, w_k\}$ is created by scanning through the training set of documents and choosing the most relevant concepts which give a good representation of the training set. Once a random query is composed using members from the set $W_0$ and the basic boolean operators, the query can be evaluated by verifying it against the training set. The boolean query is applied to each of the document in the training set, and the query predicts the document as relevant or irrelevant. The number of correct relevant or irrelevant predictions leads to the fitness for the query. The algorithm searches for those queries which provide the maximum number of correct predictions. Degeneracy often exists, as there is a possibility of more than one query producing the same results and therefore having the same fitness.

\subsection{Producing New Queries}\label{sec:produce}
\begin{figure}[hbt]
\begin{center}
\epsfig{file=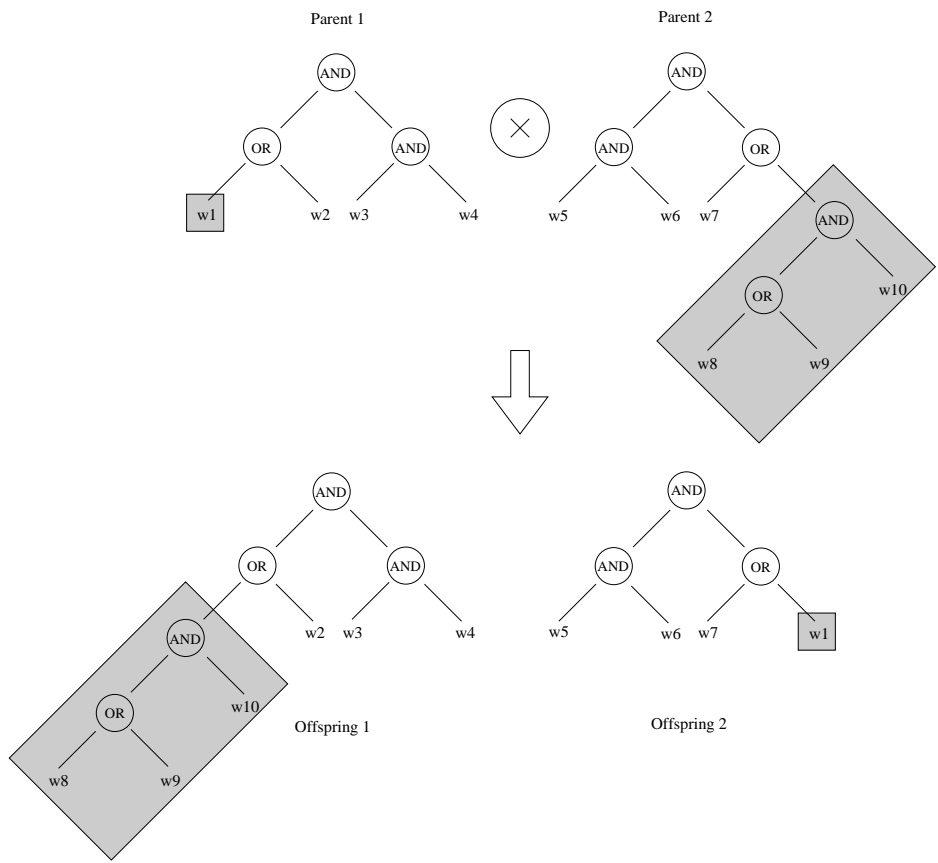,width=0.90\linewidth} 
\caption{Crossover}
\label{fig:crossover}
\end{center}
\end{figure}

New queries or offsprings are produced from the parent queries by means of crossover and mutation. A crossover method is chosen such that two parents result in two offsprings. The crossover is performed by randomly choosing a crossover point in each parent tree. Once the crossover points are chosen, the offsprings are created by swapping the subtree rooted at the crossover point of one parent with the subtree rooted at the crossover point of the other parent. Figure~\ref{fig:crossover} shows two parents and the crossover operation. The subtrees to be swapped are shown shaded in the figure. Swapping the two shaded subtrees produce the offsprings. 

\begin{figure}[hbt]
\begin{center}
\epsfig{file=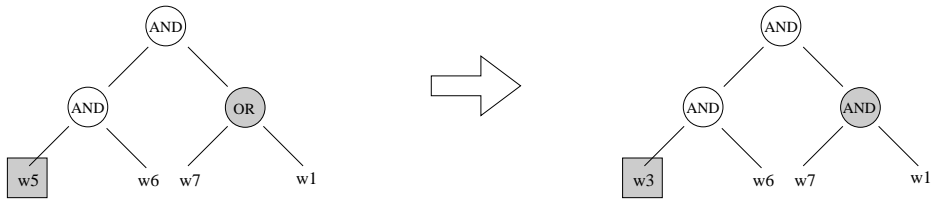,width=0.90\linewidth} 
\caption{Mutation}
\label{fig:mutation}
\end{center}
\end{figure}

Once the crossover operation is performed and the offsprings are produced, they undergo a mutation operation. A point mutation operation has been used where each node is considered in turn, and with a particular probability the primitive stored at the node is replaced with another randomly chosen primitive of the same arity \footnote{Arity means the number of arguments a function can take. In a query, a $NOT$ gate cannot be mutated with an $OR$ or $AND$ gate as $NOT$ takes a single argument as input and on the other hand $AND$ and $OR$ take two arguments as input.}. The mutation operation has been shown in Figure~\ref{fig:mutation} for the second offspring produced from crossover. Making a choice based on a mutation probability, the nodes with primitive $w_1$ and $OR$ get chosen. $w_1$ is replaced by a random member from the set, $\{w_1,w_2,\ldots,w_{10}\}$ and $OR$ is replaced by a random member from the set $\{OR, AND\}$. The crossover and mutation operation together produce the final members which compete with other members to enter the population based on their fitness. 

\subsection{Algorithm Description}\label{sec:algodesc}
\begin{figure}[htp]
\begin{center}
\epsfig{file=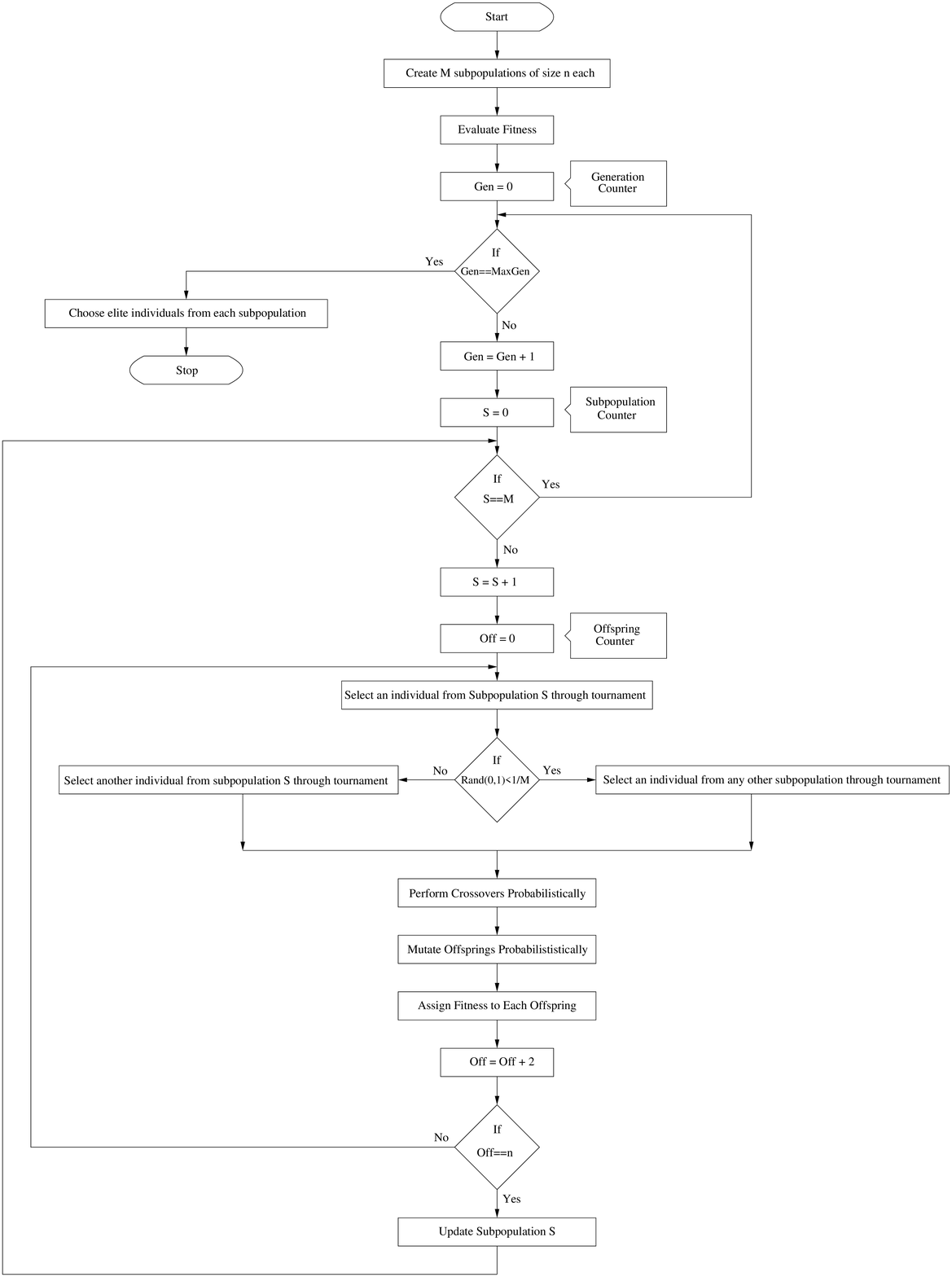,width=1.10\linewidth} 
\caption{Flowchart for GP algorithm.}
\label{fig:gp}
\end{center}
\end{figure}

The proposed algorithm follows the framework of a general evolutionary algorithm. Instead of having a single population, the algorithm maintains multiple sub-populations which interact with each other during the optimization run. The algorithm terminates when the prescribed number of generations are completed. At the end of the optimization run, the algorithm provides elites from each of the subpopulations as final solutions. These elites are expected to represent different niches in the search space. Each elite represents a Wiki-query which participates in the formation of a Wiki-ES rule. Multiple queries are accepted as solutions from the algorithm, as we do not wish to rely on a single query. For any document, output of each query is taken into account through the voting function and the decision for relevance or irrelevance is made. A flowchart for the proposed genetic programming algorithm has been presented in Figure~\ref{fig:gp}. In the following, we also discuss a stepwise procedure for implementing the algorithm.

\begin{enumerate}
\item Initialize $M$ different sub-populations randomly. Each sub-population contains $n$ number of individuals. It is noteworthy that the choice of $M$ determines the number of Wiki-queries participating in the Wiki-ESR rule, i.e. $M=|A|$ in the Definition~\ref{def:wikiESR}.
\item Assign fitness to all the initialized individuals.
\item Initialise a generation counter $Gen = 0$. 
\item If $Gen$ is less than maximum number of prescribed generations then go to Step 5, otherwise go to Step 16
\item Increment the generation counter by $1$, $Gen = Gen + 1$
\item Initialise a sub-population counter $S = 0$.
\item If $S$ is less than number of sub-populations $M$ then go to Step 8, otherwise go to Step 4
\item Increment the sub-population counter by $1$, $S = S + 1$.

\item Initialise an offspring counter $Off = 0$.
\item Choose two individuals randomly from sub-population $S$, perform a tournament and choose the better individual as one of the members for crossover.
\item Generate a random number between $0$ and $1$. If the value is less than $1/M$, then choose two individuals randomly from subpopulation other than $S$, otherwise choose two individuals randomly from the subpopulation $S$. Perform a tournament and choose the winner as the other member for crossover.
\item Perform crossover with a crossover probability $p_c$. This produces two offsprings.
\item Mutate the offsprings with a mutation probability $p_m$.
\item Increment the offspring counter by $2$, $Off = Off + 2$.
\item If offspring count, $Off$ is equal to $n$, then combine the offsprings and the individuals from the sub-population $S$ into a pool. Choose the $n$ best members from the pool, copy it into the subpopulation $S$ and go to Step 7. If offspring count, $Off$ is less than $n$ then go to Step 10
\item Choose the best members from each sub-population as final solutions.
\end{enumerate}

\subsection{Formation of Wiki-ES rules}\label{sec:formation}
As already mentioned, the suggested GP algorithm produces multiple queries as its output. If the number of sub-populations in the algorithm is chosen as $M$, then the number of final queries are also $M$ in number. Given a document, each query suggests it as either relevant or irrelevant. However, we do not want to rely on a single query, rather wish to take a weighted contribution of each of the queries before making a final decision. Let each of the query be represented by $q_i : i \in \{1,2\ldots,M\}$ and the associated fitness be represented by $F_i : i \in \{1,2\ldots,M\}$. For any given document $d$, if we need to decide whether it is relevant or irrelevant, output of each of the query is considered. Let the output of each query for the document $d$ be $b_i : i \in \{1,2\ldots,M\}$, where $b_i$ is either $0$ or $1$. Now a weighted contribution of the queries is accounted in the following metric $\mu$:
\begin{equation}
\mu = \frac{\sum_{i=1}^{M} F_i b_i}{\sum_{i=1}^{M} F_i}
\end{equation}
If the value of the metric $\mu$ is greater than $0.5$ then the document is considered relevant, otherwise it is considered irrelevant. Using this weighted contribution, the information from various niches are taken into account and overfitting of a query to the training document set is also avoided.

\section{Experiment and results}\label{sec:experiment}

To demonstrate the benefits of using Wiki-ES rules, we evaluate the system by using the topics in TREC-11 corpus. The experiment is structured as follows. First, we begin with description of the data set in Section~\ref{sec:data}, which is followed in Section~\ref{sec:system} by an account on the software components used to implement the Wiki-ES system. The parameter setup of the GP algorithm is outlined in Section~\ref{sec:param}. The results from the comparison of Wiki-ES against competing algorithms are presented in Section~\ref{sec:results}. In particular, we illustrate the benefits of using Wikipedia-concepts for query learning by benchmarking the performance of Wiki-ES against a corresponding term-based model.

\subsection{Data}\label{sec:data}

The documents included in TREC-11 corpus are Reuters RCV1 news stories from years 1996-1997. The data is partitioned into a training set (items dated between 1996-08-20 to 1996-09-30) and a test set (remainder of the collection). The training and test set are further divided into 100 topic-specific subsets. All 100 TREC-11 topics (numbered R101-R200) are used in the experiment. In this paper, only the initial training data is used, while the relevance statements available for adaptive learning are not utilized. Also none of the information in the separately available topic description file is used.

\begin{figure}[htp]
\begin{center}
\epsfig{file=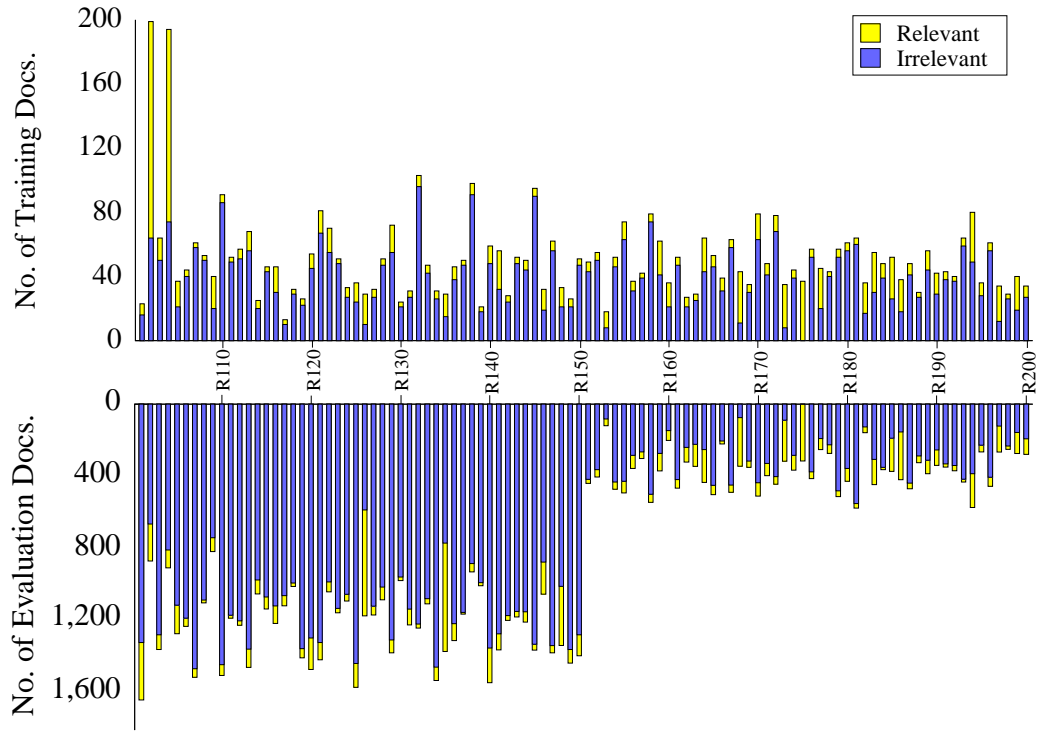,width=1\linewidth} 
\caption{The number of relevant / irrelevant documents in TREC-11 topics.}
\label{fig:data}
\end{center}
\end{figure}

Given that query learning techniques tend to be highly dependent on the quality and amount of training data, it is worthwhile to take a closer look at the data available for the 100 TREC-11 topics. Figure~\ref{fig:data} shows two histograms displaying the number of training and evaluation documents for each topic. To describe how data sets are balanced between relevant and irrelevant documents, the frequency bars are split to reflect their proportions in both data sets. On average there are 12 relevant and 39 irrelevant document examples in the training data, and 90 relevant and 713 irrelevant in the evaluation set. However, the variation between topics is quite drastic, especially in the evaluation set. As it can be seen from the histogram, the first 50 topics have a large evaluation set as compared to the remaining topics. It can also be seen that some topics are highly imbalanced, in the sense that there is only a handful of relevant documents for hundreds of irrelevant items, e.g. in the case of topic R137 less than 1\%  of the documents are relevant in the evaluation set. Then on the other extreme, a few topics (e.g. R175) are very loosely defined with majority of the documents being relevant. When considering the performance of the Wiki-ES model, as well as the benchmarks, both the quantity and balance of training data play important roles. In general, topics with relatively large proportion of relevant examples in the training data fare better than the ones with very few relevant items. The topics with few relevant documents provide good test-cases for evaluating the efficacy of the algorithm.

\subsection{System description}\label{sec:system}

The system used in the experiment was implemented using Java software on top of the GATE platform, which provides tools for standard document preprocessing tasks. The other software components used in the implementation and evaluation of Wiki-ES framework are described as follows:
\begin{itemize}
\item Wikipedia-model: The Wikipedia-based content model was built using the WikipediaMiner published by Milne et al.~\cite{milne09}, which was suitably modified and integrated into our framework. 

\item NER: The named-entity recognition task was carried out using a Conditional Random Field (CRF) classifier proposed by Finkel et al.~\cite{finkel05}. 

\item Genetic programming: The co-evolutionary GP algorithm described in Section~\ref{sec:gpalgorithm} was implemented using the JGAP toolbox provided by Meffert et al.~\cite{meffert10}.

\item Classifiers: The classifiers (SVM and C4.5) used as benchmarks in the experiment, where implemented using Weka~\cite{hall09} through Java-ML~\cite{abeel09} package.

\end{itemize}

\subsection{Parameter setting}\label{sec:param}

The GP procedure used in the paper has the usual genetic programming parameters like population size, crossover probability, mutation probability, etc. The parameter setting used in this experiment is given in Table~\ref{tab:params}.
\begin{table}[h]
\caption{GP parameters}\label{tab:params}
\centering
\begin{scriptsize}
\begin{tabular}{| l | c |} 
\hline
Parameter name & Value \\
\hline
Number of generations, $G$ & 250 \\
Number of sub-populations, $M$ & 10 \\
Sub-population size, $N$ & 100 \\
Crossover probability, $p_c$ & 0.9 \\
Mutation probability, $p_m$ & 0.9 \\
Initial tree depth & 4 \\
Maximum crossover depth & 8 \\
\hline
\end{tabular}
\end{scriptsize}
\end{table}

In addition to the general GP parameters, we have used 15 as the maximum size for the terminal set while constructing query trees. That is, when building the queries, the maximum number of different Wikipedia-concepts that could appear in a single Wiki-query was limited to 15. The choice of Wikipedia-concepts for each topic was carried out by selecting the ones that appear most frequently in the relevant training documents.

\subsection{Results}\label{sec:results}

In this section, we present the results from two experiments carried out using TREC-11 data. The first experiment, discussed in Section~\ref{sec:experiment1}, examines the importance of using Wikipedia-concepts in Wiki-ES rules by comparing them against the results obtained by running the same algorithm with bag-of-words document model. By using the bag-of-words profile in the competing model we get an effective comparison against the established IQBE-paradigm. The second experiment, presented in Section~\ref{sec:experiment2}, evaluates the benefits of Wiki-ES model in comparison to the well-known classification models based on Support Vector Machines (SVM) and the decision-tree algorithm C4.5. As performance measures, we have used F-score, precision and recall which are defined as in~\ref{def:fitness}.

\subsubsection{Experiment 1: Effect of Wikipedia semantics}\label{sec:experiment1}

Given that the main contribution of the Wiki-ES framework is the integration of Wikipedia's knowledge into the query learning problem, the first question to ask is how much the retrieval results have been improved by the infusion of the semantic information. In order to quantify the effect, we consider an experiment where the co-evolutionary GP-algorithm is run with two alternative content models: the Wikipedia-based model and the bag-of-words model. This allows us to eliminate the effect of the algorithm and focus on the improvement following from the concept-based representation of documents and queries. 

The key performance measures are summarized in Table~\ref{tab:table1}, where Token-GP refers to the model using the bag-of-words representation. The results are computed as averages across all 100 topics. A direct comparison shows that Wiki-ES yields an improvement of 62\% in F-score when compared with the Token-GP model. Interestingly, when comparing the results with respect to precision and recall, we find that most of the reported difference in F-score is due to better recall of Wiki-ES, while precision and accuracy are roughly the same. After all, recognizing the way how the concept-relatedness measure is utilised in the evaluation of Wiki-queries, the outcome was anticipated due to the ability of Wiki-queries to match such documents as well which contain a closely related concept that would have been ignored by a word based search. Whereas in the case of Token-GP based rules, it is required that the words in the query expressions are directly detected, which is likely to weaken their ability to match relevant documents.

\begin{table}[hbt]
\caption{Results for Wiki-ES and Token-GP Algorithms}
\label{tab:table1}
\begin{center}
\begin{scriptsize}
\begin{tabular}{|c|c|c|c|c|} \hline
Algorithm    &    F-Score    &    Precision    &    Recall    &    Accuracy    \\ \hline
Wiki-ES    &    0.4218    &    0.4104    &    0.5200    &    0.8436    \\ \hline
Token-GP    &    0.2596    &    0.4002    &    0.2925    &    0.8466    \\ \hline
\end{tabular}
\end{scriptsize}
\end{center}

\end{table}

\begin{figure}[htp]
\begin{minipage}[t]{0.43\linewidth}
\begin{center}
\epsfig{file=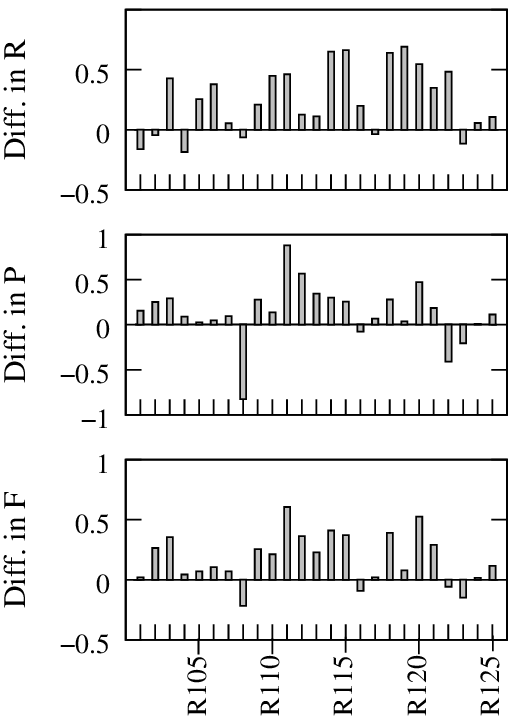,width=\linewidth}
\end{center}
\end{minipage}\hfill
\begin{minipage}[t]{0.43\linewidth}
\begin{center}
\epsfig{file=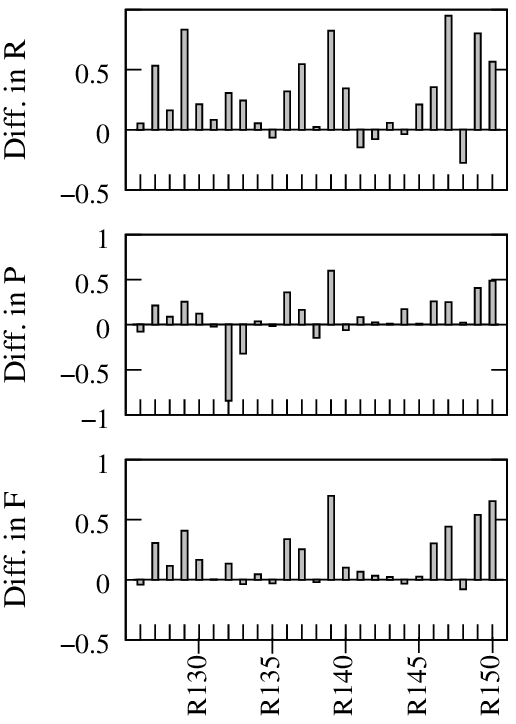,width=\linewidth}
\end{center}
\end{minipage}
\caption{Differences in F-score (F), Precision (P), and Recall (R) between models Wiki-ES and Token-GP for topics R101-R150.}\label{fig:comparison1}
\end{figure}

\begin{figure}[htp]
\begin{minipage}[t]{0.43\linewidth}
\begin{center}
\epsfig{file=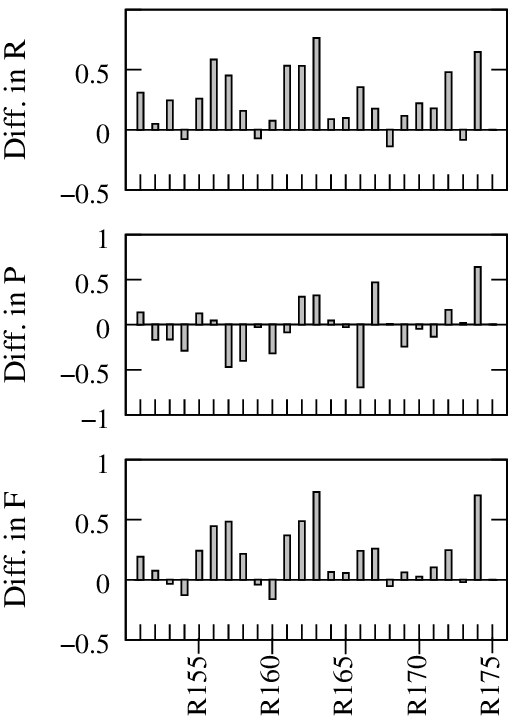,width=\linewidth}
\end{center}
\end{minipage}\hfill
\begin{minipage}[t]{0.43\linewidth}
\begin{center}
\epsfig{file=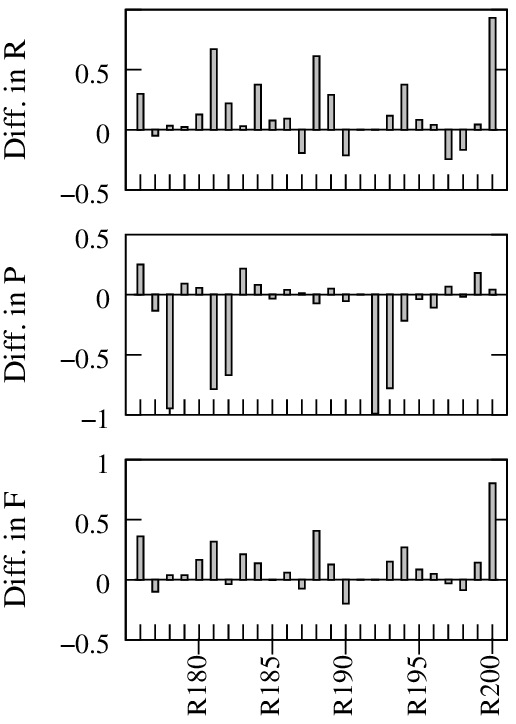,width=\linewidth}
\end{center}
\end{minipage}
\caption{Difference in F-score (F), Precision (P), and Recall (R) between models Wiki-ES and Token-GP for topics R151-R200.}\label{fig:comparison2}
\end{figure}

To provide a better idea on the differences in F-score for the two algorithms across the individual topics, Figures~\ref{fig:comparison1} and~\ref{fig:comparison2} show the difference for Wiki-ES minus Token-GP. Positive bars in the figures indicate the topics where the use of Wikipedia's semantics has been beneficial in terms of F-score, recall and precision. The reason for splitting the evaluation into subfigures stems from the characteristics of the topics. The first half of the dataset (R101-R150; Figure~\ref{fig:comparison1}) represents topics where the individual query expressions participating in the Wiki-ES rules tend to have more complicated structures. In particular, they commonly feature conditions that would require the use of NOT-gate to construct the query expressions. For example, in topic R120, we are looking for documents on deaths of mine workers where the death has occurred due to a mining accident and is not related to an ethnic clash between miners. Overall, we find that the various constraints involved in the first 50 topics make them tougher for both models.  However, when comparing the performance differences, it appears that it is exactly these difficult topics where the Wikipedia-based approach has the largest edge over Token-GP. For topics R101-R125 the average percentage improvement in F-score is 91.37\% and 82.51\% for topics R126-R150 in favor of Wiki-ES, which are both considerably larger than the improvement across all of the topics.

Also the results reported for the remaining topics (R151-R200) show that the use of Wikipedia-concepts has improved the F-scores substantially; see Figure~\ref{fig:comparison2}. However, the average percentage difference in F-score is 54.57\% for topics R151-R175 and 38.57\% for R176-R200. This suggests that although both models achieve higher F-scores than previously, the difference in their performance has essentially narrowed down on these simpler topics. When examining Figure~\ref{fig:comparison2}, we observe that even Token-GP has often performed well in terms of precision for these last 50 topics. However, Wiki-ES is still clearly outperforming in terms of improved recalls. 

To summarize, the experiment lends support for two conclusions. First, the use of Wikipedia's concept information appears to have a substantial effect on the performance of the Wiki-ES framework. The improvement stems from the ability of the rules to achieve higher recalls without losing too much precision. Second, based on the current results, it turns out that Wiki-ES's strengths are best pronounced when the query learning problem includes strict constraints that the system should be able to figure out. This combined with narrow topic definitions and meager supply of relevant documents are conditions that characterize the use-cases where the Wiki-ES rules achieve considerably better overall performance than token-based rules.

\subsubsection{Experiment 2: Comparison with classification models}\label{sec:experiment2}

The purpose of the second experiment is to compare the performance of Wiki-ES model against two well-known classification algorithms, SVM and C4.5. In order to also evaluate the effect of feature selection, the benchmark algorithms are trained using both token-based (bag-of-words) document representations and wiki-based document model. The support vector algorithms are referred to as Token-SVM and Wiki-SVM, and the decision-tree algorithms are denoted by Token-C4.5 and Wiki-C4.5, respectively.

The results are summarized in Table~\ref{tab:table2} where key performance measures are reported for each of the 5 models. A general comparison of the models suggests that the Wiki-ES framework consistently outperforms its benchmarks in terms of F-score. Once again, the primary cause for the performance advantage appears to be the improved recall of Wiki-ES rules. Whereas SVM-based models appear to yield better results if only precision would be considered. However, the recalls of Token-SVM and Wiki-SVM are quite poor, which leads to an overall modest performance. The differences in accuracies are relatively small for all of the models. 
\begin{table}[hbt]
\caption{Results for Wiki-ES, Token-C4.5, Token-SVM, Wiki-C4.5 and Wiki-SVM}
\label{tab:table2}
\begin{center}
\begin{scriptsize}
\begin{tabular}{|c|c|c|c|c|} \hline
Algorithm    &    F-Score    &    Precision    &    Recall    &    Accuracy    \\ \hline
Wiki-ES    &    0.4218    &    0.4104    &    0.5200    &    0.8436    \\ \hline
Token-C4.5    &    0.2849    &    0.2770    &    0.3730    &    0.8048    \\ \hline
Token-SVM    &    0.2215    &    0.5755    &    0.2098    &    0.8863    \\ \hline
Wiki-C4.5    &    0.3150    &    0.3478    &    0.3678    &    0.8386    \\ \hline
Wiki-SVM    &    0.2530    &    0.5649    &    0.2290    &    0.8868    \\ \hline
\end{tabular}
\end{scriptsize}
\end{center}

\end{table}

\begin{table}[hbt]
\caption{Performance matrix showing the performance of each algorithm when compared with the other algorithms. The comparison is computed as the relative difference in F-scores, $100\times(F_{algo1}-F_{algo2})/F_{algo2}$, where $F_{algo1}$ is the average F-score of the algorithm in the column and $F_{algo2}$ is the average F-score of the algorithm in the row.
}
\begin{scriptsize}

\label{tab:table3}

\begin{center}
\begin{tabular}{|c|c|c|c|c|c|c|} \hline
Algorithm    &    Wiki-ES    &    Token-GP    &    Token-C4.5    &    Token-SVM    &    Wiki-C4.5    &    Wiki-SVM    \\ \hline
Wiki-ES    &    0\%    &    -    &    -    &    -    &    -    &    -    \\ \hline
Token-GP    &    62.48\%    &    0.00\%    &    -    &    -    &    -    &    -    \\ \hline
Token-C4.5    &    48.07\%    &    -8.87\%    &    0.00\%    &    -    &    -    &    -    \\ \hline
Token-SVM    &    90.43\%    &    17.21\%    &    28.61\%    &    0.00\%    &    -    &    -    \\ \hline
Wiki-C4.5    &    33.91\%    &    -17.58\%    &    -9.56\%    &    -29.68\%    &    0.00\%    &    -19.67\%    \\ \hline
Wiki-SVM    &    66.69\%    &    2.60\%    &    12.58\%    &    -12.46\%    &    24.48\%    &    0.00\%    \\ \hline
\end{tabular}
\end{center}
\end{scriptsize}
\end{table}

Finally, to consider the effect of training data on the benchmark algorithms, we have computed relative differences in F-scores between each pair of models. The results are presented in Table~\ref{tab:table3}. For the sake of completeness Token-GP is also included in the comparison. A quick overview suggests the following observations. First of all, we find that the use of Wikipedia-concepts in document models had a positive effect on the results for all the algorithms. However, there is a substantial difference in the size of the effects. The effect of Wikipedia-concepts is large between Wiki-ES and Token-GP, but the corresponding comparisons for pairs Token-SVM vs Wiki-SVM and Token-C4.5 vs Wiki-C4.5 show only modest improvements. This is best explained by the fact that SVM and C4.5 based algorithms are not able to use concept-relatedness information while classifying documents into relevant or irrelevant. As discussed in Section~\ref{sec:querymodel}, it is the evaluation stage of Wiki-ES rules that makes them substantially different from classical approaches. Therefore, we can conclude that it is not only the Wikipedia-based document profile which makes Wiki-ES a powerful technique, but also the way the Wiki-ES rules utilise Wikipedia's concept-relatedness information while matching documents.

\section{Conclusions}\label{sec:conclusions}

The purpose of any automated query learning system is to help the user define a query that finds the items relevant to her topic. A plethora of studies exist in this direction which have been discussed in the paper. We have also discussed that the conventional frameworks lack the the concept level information contained in a word or token. Some studies have made the use of intelligent systems, still it has been difficult for them to significantly improve the performance of the information retrieval frameworks. This suggests that all these information retrieval systems inherently lack an important feature as they do not utilize the concept based information which prevents the improvement beyond a certain point.

The proposition of accessing concept level information through Wikipedia, made in this paper, provides a simple and fast technique to ingress the human-and-society level information into an information retrieval system. Wikipedia is a free and universally available database of information, which is frequently updated by the Wikipedia-community. This saves the cost and time required to maintain any such encyclopedia which justifies our choice of using Wikipedia. The implementation of Wikipedia semantics in constructing a query has produced significant improvement in results. To provide a justification for the generality of the suggestion for all the existing information retrieval systems, the idea has been implemented on two other existing frameworks leading to an improved performance. Hybridizing the concept-based-query idea with an intelligent system, genetic programming in this case, is able to produce results substantially better than what has been reported earlier. From the results it has also been observed, that the Wiki-ES framework is able to perform much better than its counterparts on the difficult topics in particular. The results obtained in the paper are promising, and the proposition made is generic, which should encourage future research in this direction. Emphasis is needed towards equipping a query with concept based knowledge, which should be able to eliminate the barriers faced by the contemporary information retrieval frameworks.



%% References


\bibliographystyle{model2-names}
\bibliography{genetic}






% \end{thebibliography}

\end{document}